\RequirePackage{fix-cm}
\documentclass[a4paper,11pt]{article}

\usepackage{authblk}
\usepackage{amsfonts,amsmath,amssymb,amsthm}
\usepackage{paralist}
\usepackage{url}
\usepackage{hyperref}
\usepackage{graphicx}
\usepackage{microtype}
\usepackage{url}
\usepackage{hyperref}
\usepackage{tikz}
\usetikzlibrary{positioning}
\usetikzlibrary{decorations.pathreplacing}

\newcommand{\N}{\mathbb{N}}

\newcommand{\F}{\mathbb{F}}

\newtheorem{problem}{Problem}

\providecommand{\keywords}[1]{\textbf{\textit{Keywords }} #1}

\begin{document}

\title{Deriving Smaller Orthogonal Arrays from Bigger Ones with Genetic Algorithms}

\author[1]{Luca Mariot}
	
\affil[1]{{\small Cyber Security Research Group, Delft University of Technology, Mekelweg 2, Delft, The Netherlands} 
	
	{\small \texttt{l.mariot@tudelft.nl}}}

\maketitle

\begin{abstract}
We consider the optimization problem of constructing a binary orthogonal array (OA) starting from a bigger one, by removing a specified amount of lines. In particular, we develop a genetic algorithm (GA) where the underlying chromosomes are constant-weight binary strings that specify the lines to be cancelled from the starting OA. Such chromosomes are then evolved through balanced crossover and mutation operators to preserve the number of ones in them. The fitness function evaluates the matrices obtained from these chromosomes by measuring their distance from satisfying the constraints of an OA smaller than the starting one. We perform a preliminary experimental validation of the proposed genetic algorithm by crafting the initial OA as a random permutation of several blocks of the basic parity-check array, thereby guaranteeing the existence of an optimal solution.
\end{abstract}

\keywords{orthogonal arrays, genetic algorithms, balanced crossover}

\section{Introduction}
\label{sec:intro}
Orthogonal Arrays (OA) are combinatorial structures that have several applications in cryptography and coding theory, such as secret sharing schemes, stream ciphers and MDS codes~\cite{hedayat12}. Formally, given a finite set $X$ of $s \in \N$ symbols, an orthogonal array of parameters $(N,k,s,t)$ is a rectangular $N\times k$ matrix with entries from $X$ such that, for any subset of $t$ columns out of $k$, each of the possible $s^t$ tuples of $t$ bits appears the same number of times $\lambda = N/s^t$. The parameters $t$ and $\lambda$ are also called respectively the \emph{strength} and the \emph{index} of the OA. Further, an OA is called \emph{simple} if there are no repeated rows in it.

Figure~\ref{fig:bin-oa-ex} depicts an example of a simple binary $OA(8,4,2,3)$ of strength $3$ and index $1$ over the set $X = \F_2$, where $\F_2 = \{0,1\}$ is the finite field with two elements.
\begin{figure}[t]
\centering
\resizebox{2cm}{!}{
\begin{tikzpicture}
[->,auto,node distance=1.5cm,
      empt node/.style={font=\sffamily,inner sep=0pt,minimum size=0pt},
      rect node/.style={rectangle,draw,font=\sffamily,minimum size=0.7cm, inner sep=0pt, outer sep=0pt}]

        %First row
	\node [rect node] (s11) {1};
        \node [rect node] (s12) [right=0cm of s11] {0};
        \node [rect node] (s13) [right=0cm of s12] {0};
        \node [rect node] (s14) [right=0cm of s13] {0};

        %Second row
	\node [rect node] (s21) [below=0cm of s11] {0};
        \node [rect node] (s22) [right=0cm of s21] {1};
        \node [rect node] (s23) [right=0cm of s22] {0};
        \node [rect node] (s24) [right=0cm of s23] {0};

        %Third row
	\node [rect node] (s31) [below=0cm of s21] {0};
        \node [rect node] (s32) [right=0cm of s31] {0};
        \node [rect node] (s33) [right=0cm of s32] {1};
        \node [rect node] (s34) [right=0cm of s33] {0};

        %Fourth row
	\node [rect node] (s41) [below=0cm of s31] {0};
        \node [rect node] (s42) [right=0cm of s41] {0};
        \node [rect node] (s43) [right=0cm of s42] {0};
        \node [rect node] (s44) [right=0cm of s43] {1};

        %Fifth row
	\node [rect node] (s51) [below=0cm of s41] {0};
        \node [rect node] (s52) [right=0cm of s51] {1};
        \node [rect node] (s53) [right=0cm of s52] {1};
        \node [rect node] (s54) [right=0cm of s53] {1};

        %Sixth row
	\node [rect node] (s61) [below=0cm of s51] {1};
        \node [rect node] (s62) [right=0cm of s61] {0};
        \node [rect node] (s63) [right=0cm of s62] {1};
        \node [rect node] (s64) [right=0cm of s63] {1};

        %Seventh row
	\node [rect node] (s71) [below=0cm of s61] {1};
        \node [rect node] (s72) [right=0cm of s71] {1};
        \node [rect node] (s73) [right=0cm of s72] {0};
        \node [rect node] (s74) [right=0cm of s73] {1};

        %Eighth row
	\node [rect node] (s81) [below=0cm of s71] {1};
        \node [rect node] (s82) [right=0cm of s81] {1};
        \node [rect node] (s83) [right=0cm of s82] {1};
        \node [rect node] (s84) [right=0cm of s83] {0};

\end{tikzpicture}
}
\caption{Example of a simple OA$(8,4,2,3)$.}
\label{fig:bin-oa-ex}
\end{figure}
As it can be seen, each subset of $3$ columns in the array of Figure~\ref{fig:bin-oa-ex} contains all $2^3=8$ $3$-bit vectors exactly once.

A very interesting application of orthogonal arrays to cryptography is for implementing \emph{masking countermeasures} to side-channel attacks (SCA)~\cite{picek15}. There, the strength parameter $t$ of the OA is related to the order of the SCA that the masking countermeasure can resist. For efficiency reasons, one needs to use an OA as small as possible, i.e. with the smallest possible number of rows $N$ for a given number of columns $k$ and strength $t$.

From a theoretical standpoint, determining the minimum number of rows admissible for an OA given its strength $t$ is an open problem. The best lower bound in this respect is Delsarte's \emph{linear programming bound}~\cite{delsarte73}, which however is not known to be tight in the general. For example, the question of finding a binary OA of $k=11$ columns and strength $t=4$ meeting Delsarte's bound of $N=96$ rows was open until very recently (see e.g. problem 2.14 in~\cite{gorodilova20}), with Wang~\cite{wang19} proving that the best known example of $N=128$ rows is the lowest number of admissible rows for that case. Consequently, there is both a theoretical and practical interest in constructing \emph{small} orthogonal arrays. Most of the approaches proposed in the literature, either based on algebraic methods or heuristic techniques, usually aim at constructing an OA from scratch.

On the other hand, in this paper we investigate the opposite direction: \emph{starting from an existing OA, try to derive a smaller one by removing some of its rows}. Clearly, the number of removed rows must be a multiple of $s^t$, to satisfy the constraint $\lambda = N/s^t$. To this end, we cast the problem in terms of optimization and design a genetic algorithm (GA) for tackling it. The GA takes in input an OA$(N,k,s,t)$ $A$ and tries to generate a smaller OA$(N',k,s,t)$ $B$ with $N'<N$ rows. In particular, $B$ is obtained by \emph{cancelling} $p=N-N'$ rows from the original OA $A$. Hence, each individual $I$ in the GA population represents a possible list of $T$ rows to be cancelled from $A$. The fitness function to be optimized, taken from~\cite{mariot18}, measures the deviation from being an OA$(N',k,s,t)$ of the matrix resulting by cancelling the rows specified by $I$ from $A$. An optimal solution is thus a list of $p$ rows that, when cancelled from $A$, results in a $N'\times k$ matrix that satisfies the definition of OA$(N',k,s,t)$, where $\lambda' = N'/s^t = \lambda - p/s^t$. The GA encodes the candidate solutions by representing them as $N$-bit strings with $t$ ones, that indicate the rows to be removed from the original matrix. Clearly, this strategy implies that the number of ones in the chromosomes must be kept constant, for which we employ a \emph{balanced} crossover operator investigated in~\cite{manzoni20} and a simple swap-based mutation operator.

As a preliminary validation, we test our GA on a very simple problem instance, which guarantees by construction the existence of an optimal solution. In particular, we start from the \emph{parity-check array} of order $t$, which is an $OA(2^t, t+1, 2, t)$ and then repeat it for $\lambda$ blocks, thereby obtaining an $OA(\lambda 2^t, t+1, 2, t)$. Then, we randomly shuffle the rows of the resulting OA. Given a new index $\lambda' < \lambda$, the task of the GA is therefore to discover a set of $(\lambda-\lambda')\cdot 2^t$ rows so that the resulting array is a shuffled repetition of $\lambda'$ blocks of the parity-check array. We perform our experiments for strength $t=4$, observing a steep increase in the difficulty of the problem already for relatively small starting sizes. This preliminary finding prompts for interesting questions to be addressed in future research, such as performing a more systematic parameter tuning phase and analyzing the associated fitness landscape, which could help in tackling also larger problem instances.

The rest of this paper is structured as follows: Section~\ref{sec:rel} briefly overviews the related work on the construction methods for OA. Section~\ref{sec:opt} formulates the search of smaller OA by removing rows as an optimization problem. Section~\ref{sec:ga} details the GA developed for this optimization problem, while Section~\ref{sec:exp} presents the preliminary experimental investigation performed to validate it. Section~\ref{sec:outro} concludes the paper and points out some directions for further research.

\section{Related Work}
\label{sec:rel}
Traditional methods for the construction of orthogonal arrays usually rely on the use of \emph{error-correcting codes} or other related algebraic methods. Indeed, the rows of an OA$(N,k,s,t)$ can be seen as codewords of an error-correcting codes, where the minimum distance is related to the strength of the OA. An excellent survey of the existing constructions based on these approaches is the book by Hedayat et al.~\cite{hedayat12}. On the other hand, the literature concerning the use of heuristic optimization techniques for constructing OA, as well as other types of combinatorial designs, is much more limited.

To the best of our knowledge, Safadi and Wang~\cite{safadi92,wang92} were the first to propose the use respectively of genetic algorithms and simulated annealing for designing \emph{mixed-level} OA, where each column can have entries from sets of different size. The authors of~\cite{rodriguez09} proposed a memetic algorithm for constructing \emph{covering arrays}, which are a generalization of OA where each $t$-uple must occur \emph{at least} $\lambda$ times in each subset of $t$ columns. Mariot et al.~\cite{mariot17} considered the problem of evolving orthogonal Latin squares (which are equivalent to OA of strength 2) defined by cellular automata rules using genetic algorithms and genetic programming (GP). Later, the same authors in~\cite{mariot18} addressed the design of binary orthogonal arrays with GA and GP.

Binary OA are also equivalent to \emph{correlation-immune} Boolean functions, which play an important role in symmetric cryptography~\cite{carlet21}. Hence, all works considering the design of correlation-immune Boolean functions with evolutionary algorithms can be seen as addressing the same problem as evolving binary OA. This includes for instance the work by Picek et al.~\cite{picek15}, where the authors employed GA and various breeds of GP to evolve correlation-immune functions of high order and minimal Hamming weight. Other works such as Mariot and Leporati~\cite{mariot15a,mariot15b} and Picek et al.~\cite{picek16a,picek16b} tackled the design of Boolean functions satisfying several cryptographic properties of interest, among which correlation-immunity, using various evolutionary and swarm intelligence methods including GA, discrete particle swarm optimization (PSO) and Cartesian GP.

\section{Removing OA Rows as an Optimization Problem}
\label{sec:opt}
We now define the task of removing rows from an initial OA to obtain a smaller one as an optimization problem. In what follows, we will denote an OA as a set of $N$ vectors over the set $X$ of length $k$ each. Indeed, the order of the rows in an OA is not important, since it does not influence the balancedness constraint. Let $A = \{r_1, \cdots, r_n\}$ be an OA$(N,k,s,t)$ where $r_i \in X^k$ for all $i \in \{1,\cdots, N\}$, and let $\lambda = N/s^t$ be the index of $A$. Given a smaller index $\lambda' < \lambda$ and a set of $p$ rows $T = \{i_1,\cdots, i_p\}$, with $i_j \in \{1,\cdots, N\}$ for all $j \in \{1,\cdots,p\}$, define the new array $B$ as follows:
\begin{equation}
\label{eq:barr}
B = A \setminus \{r_{i_1},\cdots, r_{i_p} \} \enspace .
\end{equation}
In other words, $B$ is the array obtained by removing the rows specified by $T$ from $A$. Although being an $N'\times k$ binary array with $N'= \lambda'\cdot s^t$, in general $B$ will not satisfy the property of an OA$(N',k,s,t)$. Therefore, we can state our optimization problem of interest as follows:
\begin{problem}
\label{prob:stat}
Let $A$ be an OA$(N,k,s,t)$ with $\lambda = N/s^t$, and let $\lambda' < \lambda$. Find a set $T = \{i_1,\cdots, i_p\}$ of $p = (\lambda-\lambda')\cdot s^t$ rows such that the array $B$ as defined in Eq.~\eqref{eq:barr} is an $OA(N',k,s,t)$, with $N' = \lambda' s^t$.
\end{problem}
To measure the fitness of a candidate solution $T$ to Problem~\ref{prob:stat}, we employ the same fitness function defined in~\cite{mariot18} for evolving binary OA. The idea is to compute the \emph{Minkowski distance} of the vector of occurrences of each $t$-uple from the vector $(\lambda',\lambda',\cdots,\lambda')$. In particular, the array $B$ resulting from the removal of the rows specified by $T$ will be an OA$(N',k,s,t)$ if and only if such distance is $0$. Therefore, the optimization objective is to \emph{minimize} the fitness function. Due to the lack of space, we refer the reader to~\cite{mariot18} for the formal definition of the fitness function. In all our experiments, we used the Minkowsky distance with exponent $2$, which basically corresponds to the Euclidean distance. 

\section{Genetic Algorithm}
\label{sec:ga}
In order to design a genetic algorithm for tackling Problem~\ref{prob:stat}, one first needs to define the chromosome encoding of the candidate solutions, which in our cases are lists of rows to be removed from the original OA. In set-theoretic terms, the most straightforward way is to use a binary vector that represents the characteristic function of the set $T = \{i_1,i_2,\cdots,i_p\}$. Thus, given the orthogonal array $A$ of $N$ rows, the chromosome $C_T$ of a candidate solution $T$ is a binary string of length $N$ whose coordinates are defined as:
\begin{equation}
\label{eq:chrom}
C_t[i] = 
\begin{cases}
1, & \textrm{ if } i \in T \enspace , \\
0, & \textrm{ otherwise}
\end{cases}
\enspace ,
\end{equation}
for all $i \in \{1,\cdots, N\}$. In particular, each chromosome has a constant \emph{Hamming weight}, that is the number of ones in it is always fixed to $p$, which corresponds to the size of the candidate solution $T$. Clearly, the constraint above raises the question of how to make the genetic algorithm preserve the number of ones in the offspring chromosomes, so that they always represent a valid list of $p$ rows to be removed from the original OA. To this end, we respectively used a \emph{balanced operator} for crossover and a \emph{swap-based operator} for mutation.

Such operators have been mostly investigated in the literature related to the optimization of Boolean functions with good cryptographic properties. Since one of the basic properties that such functions must satisfy to be used in a stream or block cipher is to be \emph{balanced} (i.e., having the same number of ones and zeros in their truth tables), several researchers investigated the design of ad-hoc crossover operator to ensure this constraint, in order to reduce the size of the search space explored by a GA. In particular, Millan et al.~\cite{millan98} proposed a crossover operator where two counters are used to keep trace of the multiplicities of zeros and ones while the offspring chromosome is being created from the parents. Once one of the counters reaches the prescribed threshold, the offspring is filled with the complementary value in the remaining positions. Other works~\cite{mariot15b,mariot17,manzoni19} considered variations of this operator over non-binary strings and for other problems. Manzoni et al.~\cite{manzoni20} performed a systematic comparison of three different balanced crossover operators, observing that the \emph{map-of-ones} operator usually performs better. For this reason, we adopted it for our GA. The idea of this crossover is to represent the two parent strings in terms of their map of ones, that is the list of positions in the strings where the ones occur. Then, the offspring map is constructed by randomly copying from the parents' maps, checking consistency to avoid that duplicate positions are inserted in the offspring. In this way, starting from two parents that have the same Hamming weight (that is, map of ones of the same length), the child chromosome is guaranteed to inherit the same weight.

Regarding the mutation phase, instead, we opted for the simple operator used in~\cite{manzoni20}, which randomly selects a pair of positions in the bitstrings holding different values and swap them.

\section{Experiments}
\label{sec:exp}
As a preliminary assessment of our GA for Problem~\ref{prob:stat}, we adopted the following experimental setup considering only the case of binary OA (hence, with $s=2$ and $X = \F_2)$. For the initial OA, we started with the \emph{parity-check} array as a basic building block, also called the zero-sum array in~\cite{hedayat12}. The parity-check array $P$ of order $t$ is defined as a $(2^t) \times (t+1)$ binary matrix where the first $t$ columns holds all $2^t$ binary vectors in $\F_2^t$ (for example, in lexicographic order). The last column, on the other hand, contains the results of the XOR of the bits in the previous columns. It is rather easy to prove that such an array $P$ is an OA$(2^t,t+1,2,t)$. Then, given the desired starting index $\lambda$, we constructed a $(\lambda2^t) \times (t+1)$ binary array by repeating $\lambda$ times the OA $P$, randomly shuffling its rows at the end. Obviously, the resulting array is an OA$(\lambda\cdot 2^t,t+1,2,t)$, which is not simple since each rows occurs exactly $\lambda$ times.

For our experiments, we considered the case of strength $t=4$, with the index $\lambda$ for the initial OA ranging between $2$ and $4$. Hence, the smallest problem instance consisted in starting from an OA$(32,5,2,4)$ ($\lambda = 2$) and finding a subset of 16 rows such that the reduced matrix is an OA$(16,5,2,4)$ ($\lambda' = 1$). In this case, the size of the search space is $\binom{32}{16} = 6.32\cdot10^8$, which in principle is still amenable to exhaustive search, but nonetheless provides an interesting case to gauge the performances of our GA. The largest instance, on the other hand, was to start from an OA$(64,5,2,4)$ ($\lambda = 4$) and find a subset of $32$ rows to erase in order to obtain an OA$(32,5,2,4)$ ($\lambda' = 2$). In this case the search space size is $\binom{64}{32} \approx 1.82\cdot 10^{18}$, which cannot be exhaustively searched.

Regarding the parameters of our GA, we drew upon those used in~\cite{mariot18}, which also targeted the construction of binary OA: steady-state selection with tournament size 3, population size of 500 individuals, mutation probability $0.2$, and a fitness budget of $100\,000$ evaluations. Finally, we repeated each experiment for $30$ independent runs to obtain statistically significant results.

Table~\ref{tab:res} reports, for each of the considered $6$ problem instances, the number of optimal solutions found over the $30$ independent runs and the median fitness value of the best solution evolved by the GA.
\begin{table}[t]
\centering
\begin{tabular}{cccc}
\hline\smallskip
$\lambda,\lambda'$ & 1 & 2 & 3 \\
\hline
\hline
2 & (24/30, 0.0) & $-$ & $-$ \\
3 & (9/30, 7.07) & (4/30, 7.07) & $-$ \\
4 & (8/30, 7.07) & (0/30, 7.07) & (8/30, 7.07) \\
\hline
\smallskip
\label{tab:res}
\end{tabular}
\caption{Number of optimal solutions and median fitness over all problem instances.}
\end{table}
It can be observed that there is a steep degradation in performances as soon as one leaves the smallest problem instance with $\lambda=2$ and $\lambda'=1$. While in this case the GA almost always converges to an OA$(16,5,2,4)$ starting from an OA$(32,5,2,4)$, the situation worsens already for $\lambda=3$ with only $9$ and $4$ optimal solutions found respectively for $\lambda' = 1$ and $\lambda' = 2$. Over the largest instance, i.e. $\lambda=4$ and $\lambda'=2$, the GA never converges to an optimal solution. It is also interesting to note that, except for the smallest problem instance, the median fitness is always the same. In fact, we found that the fitness distribution for the final best individual is bi-modal over all instances tackled by our GA, with the only observed values being $0.0$ and $7.07$.

\section{Conclusions}
\label{sec:outro}
In this paper, we proposed a genetic algorithm to evolve orthogonal arrays with small index starting from bigger ones. The basic idea is to represent a candidate solution as a set of rows to be removed from the original arrays, which are represented by bitstrings with a constant number of ones. The GA then evolves such bitstrings by using ad-hoc crossover and mutation operators that preserve the Hamming weight of the strings.

The preliminary results gathered in our investigation show that this optimization problem seems to be extremely difficult for GA: indeed, already for small instances such as removing 32 rows from an OA$(64,5,2,4)$, our GA was not able to produce any optimal solution. Also, we noticed that all obtained distributions of the best fitness are bi-modal. These empirical observations indicate that, in future research, two directions should be particularly considered: first, analyzing the fitness landscape of this optimization problem might help in understanding why the GA gets stuck in the local optima with fitness value 7.07. As a matter of fact, it would be interesting to analyze the solutions to which our GA converges, to verify whether it is always the same, or if several different solutions with the same sub-optimal fitness exist. Second, we suspect that a more systematic parameter tuning phase could benefit the performances of our GA on the larger problem instances.

\bibliographystyle{abbrv}
\bibliography{bibliography}

\end{document}